\documentclass[letterpaper, 10 pt, conference]{ieeeconf}

\IEEEoverridecommandlockouts %

\usepackage{amsmath,amsfonts,amssymb}
\usepackage{amsfonts} %
\usepackage{mathrsfs} %
\usepackage{bbold} %
\usepackage{pgfplots}
\usepackage[ruled, vlined, linesnumbered]{algorithm2e}
\usepackage[tight]{subfigure}
\usepackage{tabularx}
\usepackage{multirow}
\usepackage{siunitx}

\usepackage[export]{adjustbox}

\graphicspath{{figures/}}

\usepackage{caption}
\usepackage{xargs} %

\makeatletter
\def\ignorecitefornumbering#1{%
	\begingroup
	\@fileswfalse
	#1%
	\endgroup
}
\makeatother
\usepackage[sorting=none, style=ieee, backend=biber, bibencoding=utf8, maxbibnames=5, mincitenames=1, maxcitenames=2, url=false, doi=false, isbn=false, citestyle=numeric-comp, natbib=true, eprint=false]{biblatex}

\usepackage{bm} %
\usepackage{lipsum}
\usepackage{glossaries}
\newacronym{fov}{FOV}{Field Of View}
\newacronym{ins}{INS}{Inertial Navigation System}
\newacronym{rtk}{RTK}{Real-Time Kinematic}
\newacronym{acc}{ACC}{Adaptive Cruise Control}
\newacronym{phev}{PHEV}{Plug-in Hybrid Electric Vehicle}
\newacronym{darpa}{DARPA}{Defense Advanced Research Projects Agency}
\newacronym{asild}{ASIL D}{Automotive Safety Integrity Level D}
\newacronym{lc}{LC}{Lane Change}
\newacronym{rtcomp}{RT comp.}{Real-Time computing system}
\newacronym{ros}{ROS}{Robot Operating System}
\newacronym{lcm}{LCM}{Lightweight Communication and Marshalling}
\newacronym{gcdc}{GCDC}{Grand Cooperative Driving Challenge}
\newacronym{v2v}{V2V}{Vehicle-to-Vehicle}
\newacronym{fog}{FOG}{Fiber Optic Gyro}
\newacronym{gnss}{GNSS}{Global Navigation Satellite System}
\newacronym{dof}{DoF}{Degrees of Freedom}
\newacronym{ipc}{IPC}{Inter Process Communication}
\newacronym{icp}{ICP}{Iterative Closest Point}
\newacronym{imu}{IMU}{Inertial Measurement Unit}
\newacronym{gps}{GPS}{Global Positioning System}
\newacronym{ransac}{RANSAC}{Random sample consensus}
\newacronym{v2x}{V2X}{Vehicle-to-Everything}
\newacronym{gmsl}{GMSL}{Gigabit Multimedia Serial Link}
\newcommand{\Secref}[1]{Section~\ref{#1}} 

\newcommand\copyrighttext{%
	\scriptsize \textcolor{blue}{\textcopyright 2019 IEEE. Personal use of this material is permitted.  Permission from IEEE must be obtained for all other uses, in any current or future media, including reprinting/republishing this material for advertising or promotional purposes, creating new collective works, for resale or redistribution to servers or lists, or reuse of any copyrighted component of this work in other works}}
\newcommand\copyrightnotice{%
	\begin{tikzpicture}[remember picture,overlay]
	\node[anchor=north,yshift=-7.5pt] at (current page.north) {\fbox{\parbox{\dimexpr\textwidth-\fboxsep-\fboxrule\relax}{\copyrighttext}}};
	\end{tikzpicture}%
}

\usepackage{balance}

\pgfplotsset{compat=newest}
\pgfplotsset{plot coordinates/math parser=false}
\usetikzlibrary{decorations.markings, positioning}

\usepackage{booktabs}
\newcommand{\ra}[1]{\renewcommand{\arraystretch}{#1}}

\newlength\figureheight 
\newlength\figurewidth 

\title{Bridging the Gap between Open Source Software and Vehicle Hardware for Autonomous Driving}

\author{Tobias Kessler$^{1}$, Julian Bernhard$^{1}$, Martin Buechel$^{1}$, Klemens Esterle$^{1}$, Patrick Hart$^{1}$, Daniel Malovetz$^{1}$, \\
Michael Truong Le$^{1}$, Frederik Diehl$^{1}$, Thomas Brunner$^{1}$ and Alois Knoll$^{2}$%
	\thanks{$^{1}$fortiss GmbH, An-Institut Technische Universit\"{a}t M\"{u}nchen, Munich, Germany}%
	\thanks{$^{2}$Chair of Robotics, Artificial Intelligence and Real-time Systems, Technische Universit\"{a}t M\"{u}nchen, Munich, Germany}%
}

\begin{document}

\maketitle
\copyrightnotice
\thispagestyle{empty}
\pagestyle{empty}

\global\csname @topnum\endcsname 0
\global\csname @botnum\endcsname 0

\newcommand{\figurename}{Fig. }

\newcommand {\vect} {\boldsymbol}
\newcommand {\matr} {\boldsymbol}

\newcommand{\state} {\vect{x}}
\newcommand{\stateSpace} {\vect{\mathcal{X}}}
\newcommand{\beliefstate} {\vect{b}}
\newcommand{\contr} {\vect{u}}
\newcommand{\contrSpace} {\vect{\mathcal{U}}}
\newcommand{\meas} {\vect{y}}
\newcommand{\procNoise}{\vect{w}}

\newcommand {\cov}  {\matr{\Sigma}}

\newcommand{\stateNoDelta}{\hat\state}
\newcommand{\contrNoDelta}{\hat\contr}
\newcommand{\procNoiseNoDelta}{\hat\procNoise}

\newcommand{\abc}[2][\empty]{%
  \ifthenelse{\equal{#1}{\empty}}
    {no opt, mand.: \textbf{#2}}
    {opt: \textbf{#1}, mand.: \textbf{#2}}
}

\newcommand {\noiseu} {\procNoise}
\newcommand {\covu} {\matr{\Sigma_{\noiseu,}}}

\newcommand {\noiseuNoDelta} {\procNoiseNoDelta}
\newcommand {\covuNoDelta} {\matr{\Sigma_{\noiseuNoDelta,}}}

\newcommand {\defnoiseu}[1][\empty]{
 \ifthenelse{\equal{#1}{\empty}}
    {\noiseu\sim N(0,\covu)}
    {\noiseu_{#1}\sim N(0,\covu_{#1})}
}

\newcommand {\defnoiseuNoDelta}[1][\empty]{
 \ifthenelse{\equal{#1}{\empty}}
    {\noiseuNoDelta\sim N(0,\covuNoDelta)}
    {\noiseuNoDelta_{#1}\sim N(0,\covuNoDelta_{#1})}
}

\newcommand {\covm} {\matr{R}}
\newcommand {\noisem} {\vect{\nu}}
\newcommand {\defnoisem}[1][\empty]{
 \ifthenelse{\equal{#1}{\empty}}
    {\noisem\sim N(0,\covm)}
    {\noisem_{#1}\sim N(0,\covm_{#1})}
}

\newcommand{\stateB}{\vect{\xi}}
\newcommand{\contrB}{\vect{\nu}}
\newcommand{\procNoiseB}{\vect{\omega}}

\newcommand{\AB}{\mathcal{A}}
\newcommand{\BB}{\mathcal{B}}
\newcommand{\WB}{\mathcal{W}}
\newcommand{\costStateB}{\mathcal{Q}}
\newcommand{\costContrB}{\mathcal{R}}
\newcommand{\covStatesB}{\mathcal{S}_{\state}}
\newcommand{\covProcNoiseB}{\mathcal{S}_{\procNoise}}

\newcommand{\FB}{\mathcal{F}}

\newcommand{\cct}{\vect{t}}
\newcommand{\ccsval}{s}
\newcommand{\ccT}{\matr{T}}
\newcommand{\ccsvec}{\vect{s}}

\newcommand{\costState}{\matr{Q}}
\newcommand{\costContr}{\matr{R}}
\newcommand{\feedbackMatrix}{\matr{K}}
\newcommand{\cost}{J}

\newcommand{\stateConstraintMatrix}{\matr{C}}
\newcommand{\stateConstraintVector}{\vect{c}}

\newcommand{\stateConstraintFunc}{c}

\newcommand{\contrConstraintMatrix}{\matr{D}}
\newcommand{\contrConstraintVector}{\vect{d}}

\newcommand{\contrConstraintFunc}{d}

\newcommand{\stateRef}{\state^{*}}
\newcommand{\contrRef}{\contr^{*}}

\newcommand{\stateDelta}{\Delta\state}
\newcommand{\contrDelta}{\Delta\contr}

\newcommand {\Comment}[1]{\textcolor{blue}{#1}}

\newcommand {\partialder}[4][\bigg]{\frac{\partial #2}{\partial #3}#1|_{#4}}
\newcommand {\partialdernoarg}[3][\bigg]{\frac{\partial #2}{\partial #3}#1}

\newcommand{\nat}{\mathbb{N}}
\newcommand{\real}{\mathbb{R}}
\newcommand{\compl}{\mathbb{C}}

\newcommand{\norm}[1]{\left\| #1 \right\|}

\newcommand{\half}{\frac{1}{2}}

\newcommand{\parenth}[1]{ \left( #1 \right) }
\newcommand{\bracket}[1]{ \left[ #1 \right] }
\newcommand{\accolade}[1]{ \left\{ #1 \right\} }
\newcommand{\pardevS}[2]{ \delta_{#1} f(#2) }
\newcommand{\pardevF}[2]{ \frac{\partial #1}{\partial #2} }

\newcommand{\vecii}[2]{\begin{pmatrix} #1 \\ #2 \end{pmatrix}}
\newcommand{\veciii}[3]{\begin{pmatrix}  #1 \\ #2 \\ #3	\end{pmatrix} }
\newcommand{\veciv}[4]{\begin{pmatrix}  #1 \\ #2 \\ #3 \\ #4	\end{pmatrix}}

\newcommand{\matii}[4]{\left[ \begin{array}[h]{cc} #1 & #2 \\ #3 & #4 \end{array} \right]}
\newcommand{\matiii}[9]{\left[ \begin{array}[h]{ccc} #1 & #2 & #3 \\ #4 & #5 & #6 \\ #7 & #8 & #9	\end{array} \right]}

\newcommand{\transp}{^{\intercal}}
\newcommand{\Reg}{$^{\textregistered}$}
\newcommand{\reg}{$^{\textregistered}$ }
\newcommand{\Tm}{\texttrademark}
\newcommand{\tm}{\texttrademark~}
\newcommand {\bsl} {$\backslash$}

\newtheorem{theorem}{Theorem}[section]
\newtheorem{lemma}[theorem]{Lemma}
\newtheorem{corollary}[theorem]{Corollary}
\newtheorem{remark}[theorem]{Remark}
\newtheorem{definition}[theorem]{Definition}
\newtheorem{equat}[theorem]{Equation}
\newtheorem{example}[theorem]{Example}
\newcommand{\insertfigure}[4]{ %
	\begin{figure}[htbp]
		\begin{center}
			\includegraphics[width=#4\textwidth]{#1}
		\end{center}
		\vspace{-0.4cm}
		\caption{#2}
		\label{#3}
	\end{figure}
}

\newcommand{\refFigure}[1]{\figurename \ref{#1}}
\newcommand{\refChapter}[1]{chapter \ref{#1}}
\newcommand{\refParagraph}[1]{paragraph \ref{#1}}
\newcommand{\refEquation}[1]{(\ref{#1})}
\newcommand{\refTable}[1]{Table \ref{#1}}
\newcommand{\refAlgorithm}[1]{Algorithm \ref{#1}}

\newcommand{\rigidTransform}[2]
{
	${}^{#2}\!\mathbf{H}_{#1}$
}

\newcommand{\code}[1]
 {\texttt{#1}}

\newcommand{\comment}[1]{\marginpar{\raggedright \noindent \footnotesize {\sl #1} }}

\newcommand{\clearemptydoublepage}{%
  \ifthenelse{\boolean{@twoside}}{\newpage{\pagestyle{empty}\cleardoublepage}}%
  {\clearpage}}

\newcommand{\etAl}{\emph{et al.}\mbox{ }}

\newcommand{\todoi}[1]{\todo[inline]{#1}}

\begin{abstract}
Although many research vehicle platforms for autonomous driving have been built in the past, hardware design, source code and lessons learned have not been made available for the next generation of demonstrators. This raises the efforts for the research community to contribute results based on real-world evaluations as engineering knowledge of building and maintaining a research vehicle is lost.
In this paper, we deliver an analysis of our approach to transferring an open source driving stack to a research vehicle. 

We put the hardware and software setup in context to other demonstrators and explain the criteria that led to our chosen hardware and software design.
Specifically, we discuss the mapping of the Apollo driving stack to the system layout of our research vehicle, \textsl{fortuna}, including communication with the actuators by a controller running on a real-time hardware platform and the integration of the sensor setup. 
With our collection of the lessons learned, we encourage a faster setup of such systems by other research groups in the future.

\end{abstract}

\IEEEpeerreviewmaketitle

\begin{figure}[t]	
	\def\svgwidth{\columnwidth}
	\footnotesize
	\includegraphics[width=\columnwidth]{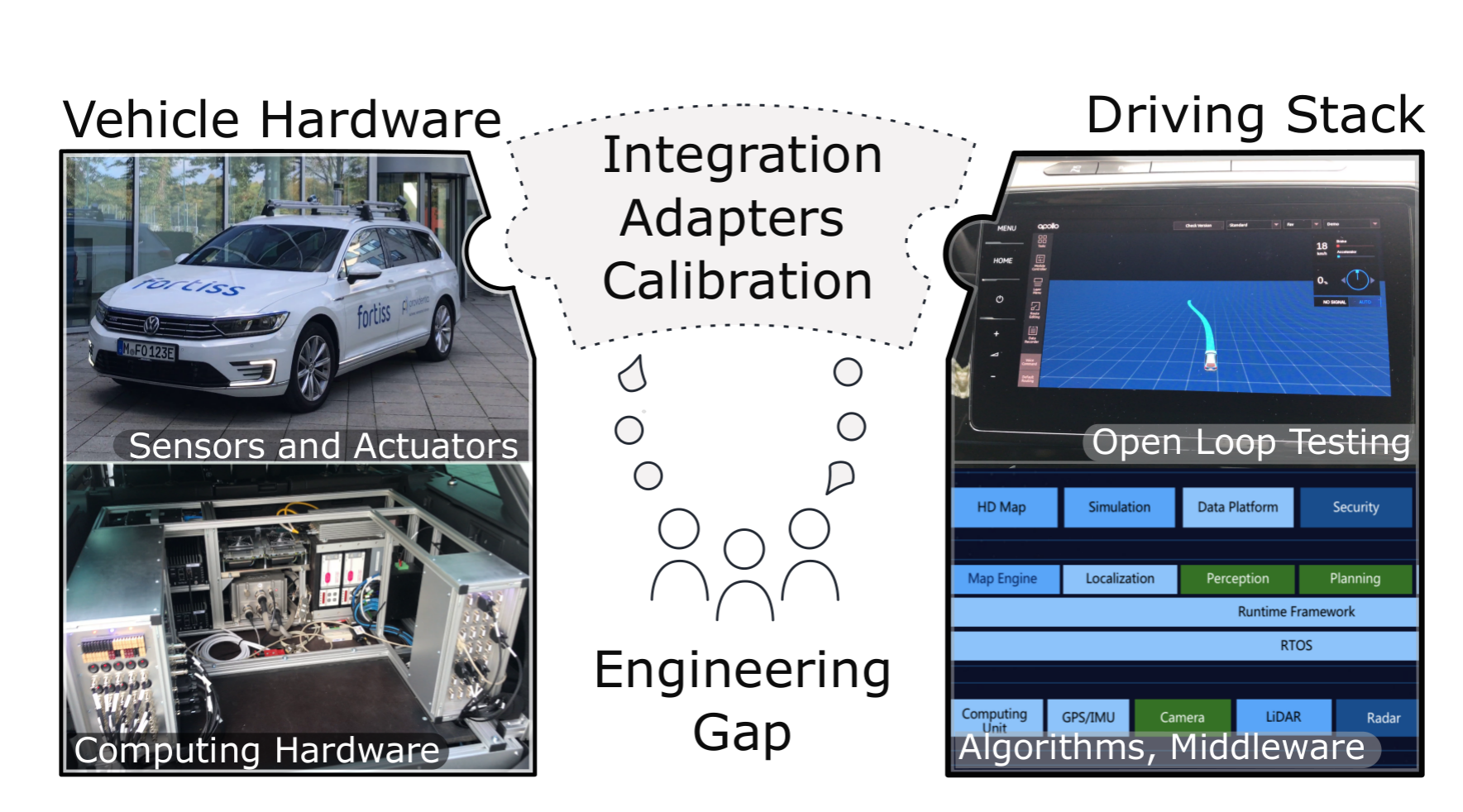}
	\caption{Autonomous driving software has to be deployed to vehicle hardware. Know-how, code, and pitfalls shall be mirrored back to hardware and software design and the research community.}
	\label{fig:fortuna}
\end{figure}

\section{Introduction}
\label{sec:introduction}

\begin{table*}[tb]
	\centering
	\vspace{0.15cm}
	\caption{Hardware comparison of relevant research vehicles from an early demonstrator, the DARPA challenges and the Apollo reference vehicle. We use $\blacklozenge$ to indicate an existing component, $\Diamond$ to indicate its nonexistence and $?$ if such information is not public.}
	\scriptsize
	\ra{1.3}
	\begin{tabular}{@{}llllllllllllllll@{}}\toprule
& \multicolumn{1}{c}{1994}  & \phantom{.} & \multicolumn{2}{c}{2007} & \phantom{.} & \multicolumn{1}{c}{2013} & \phantom{.} & \multicolumn{1}{c}{2015} & \phantom{.} & \multicolumn{2}{c}{2016} & \phantom{.} & \multicolumn{1}{c}{2018} & \phantom{.}  & \multicolumn{1}{c}{2019}\\
\cmidrule{2-2} \cmidrule{4-5} \cmidrule{7-7} \cmidrule{9-9} \cmidrule{11-12} \cmidrule{14-14} \cmidrule{16-16}
& VaMP \ignorecitefornumbering{\cite{Thomanek1996}} && Junior \ignorecitefornumbering{\cite{Montemerlo2008}} & Boss \ignorecitefornumbering{\cite{Urmson2008}} && Bertha \ignorecitefornumbering{\cite{Ziegler2014a}} && RACE \ignorecitefornumbering{\cite{Buechel2015}} && Halmstad \ignorecitefornumbering{\cite{Aramrattana2018}} & Bertha \ignorecitefornumbering{\cite{Tas2018}} && Apollo \ignorecitefornumbering{\cite{Baidu2017}} && \textsl{fortuna}\\ \midrule
Camera & $\blacklozenge$ front/rear&& $\Diamond$ & $\blacklozenge$ front && $\blacklozenge$ stereo && $\blacklozenge$ front && $\Diamond$ & $\blacklozenge$ stereo/360$^\circ$&& $\blacklozenge$ front && $\blacklozenge$ 360$^\circ$ \\
Lidar & $\Diamond$ && $\blacklozenge$ 64L& $\blacklozenge$ && $\Diamond$ && $\blacklozenge$ 4L && $\Diamond$ & $\blacklozenge$ 4L && $\blacklozenge$ 64L && $\blacklozenge$ 32L\\
Radar & $\Diamond$ && $\blacklozenge$ & $\blacklozenge$ && $\blacklozenge$ && $\blacklozenge$ && $\blacklozenge$ (series) & $\blacklozenge$ && $\blacklozenge$ && $\blacklozenge$\\
GPS & $\Diamond$ && $\blacklozenge$ & $\blacklozenge$ && $\blacklozenge$ && $\blacklozenge$ && $\blacklozenge$ rtk & $\blacklozenge$ rtk && $\blacklozenge$ rtk && $\blacklozenge$ rtk\\
INS & $\blacklozenge$ && $\blacklozenge$ & $\blacklozenge$ && ? && $\blacklozenge$ && $\blacklozenge$ & $\blacklozenge$ && $\blacklozenge$ && $\blacklozenge$\\
\\
RT comp. & $\blacklozenge$ && ? & $\Diamond$ && ? && $\blacklozenge$ && $\blacklozenge$ & $\blacklozenge$ && $\Diamond$ && $\blacklozenge$\\
PC & $\blacklozenge$ && $\blacklozenge$ & $\blacklozenge$ && ? && $\blacklozenge$ && $\blacklozenge$ & $\blacklozenge$ && $\blacklozenge$ && $\blacklozenge$\\
GPU & $\Diamond$ && $\Diamond$ & $\Diamond$ && ? && $\Diamond$ && $\Diamond$ & $\blacklozenge$ && $\blacklozenge$ && $\blacklozenge$\\
\bottomrule
\end{tabular}
	\label{tab:hardware}
\end{table*}

The participants \cite{Montemerlo2008, Urmson2008, Kammel2008, VonHundelshausen2008} of the \gls{darpa} Urban Challenge demonstrated that full-sized research vehicles foster identification of essential research problems while testing and verifying algorithms in real-world scenarios. To fully leverage the potential of the research community in the field of autonomous driving, academic institutions should not only rely on OEMs for real-world experiments but evaluate algorithms on their own research platforms.

Despite the rich history of past research vehicles starting in the early 90s (see \Secref{sec:related_work:other}), the source code has rarely been made open source. Apart from a collaborative approach by \citet{Levinson2011} to summarize lessons learned on the algorithm side, only sparse knowledge has flown back to the community regarding how to build up an architecture including the inevitable pitfalls one will face. The fully autonomous Bertha Benz Memorial drive, for example, has been a celebrated milestone with significant scientific contributions \cite{Ziegler2014a}. 
However, as we outline in \refTable{tab:hardware}, the hardware configuration, the sensor set, and others have not been published to protect the intellectual property of the OEM. The research community thus often remains unable to reproduce and verify new algorithms as the technical and organizational hurdles to build up and operate a research platform are very high.

Only recently, open source driving stacks got attention, potentially enabling research groups around the world to solve real-world problems. 
\textit{Driving stack} generally denotes the set of all software components that are necessary for fully autonomous driving.
A prominent example is \textit{Apollo}, an open source project funded and operated by Baidu \cite{Baidu2017}. 
A joint project of various Japanese universities has produced the open-source stack \textit{Autoware} \cite{Kato2018} that claims similar capabilities as Apollo. 
Both projects expect deployment on a specific exclusive set of supported hardware architectures.

However, transferring a software stack from a specific vehicle hardware architecture, including sensor setup, actuation interfaces, and computational hardware, to another is challenging due to missing standardized hardware configurations for autonomous vehicles. 
As an in-depth discussion of the architectures is often neglected in scientific publications, engineering knowledge is lost from one generation of researchers to another.

We provide a systematic analysis of the steps we applied to integrate the Apollo Driving Stack on our research vehicle with a detailed presentation of the lessons learned (see Fig. \ref{fig:fortuna}). 
We designed the vehicle as a research prototype for the development of cognitive systems and autonomous driving function prototyping.
The presented modules bridge the gap between the hardware and driving stack. We specifically contribute
\begin{itemize}
	\item a detailed discussion on how and for which reasons we modified a production vehicle equipped with state-of-the-art sensors and the access to control lateral and longitudinal motion, starting from a discussion on other research demonstrators,
	\item how we used and adopted the open-source Apollo stack,
	\item the lessons learned as a guideline for the research community.
\end{itemize}

The paper is structured as follows: First, we compare our platform to previous research vehicles. Then, we present the challenges of transferring the Apollo stack to our vehicle. In the end, we detail the sensor and control architecture setups and algorithms adapted to run with Apollo on \textsl{fortuna}.

\section{Relation to Other Research Platforms}
\label{sec:related_work}

Our research vehicle \textsl{fortuna} shall serve as a platform to study variants of automated driving algorithms. 
Our institute has previously studied architectural aspects with the vehicle demonstrator described in \Secref{sec:race}.
Various other autonomous vehicle research platforms have already been constructed. 
In \Secref{sec:related_work:other}, we present a selected set of vehicles.

\subsection{\textsl{fortuna} and the RACE Demonstrator in Contrast}
\label{sec:race}
The project RACE demonstrated a robust and reliant centralized electronic system architecture and a runtime environment which supports the integration of mixed-criticality components up to \gls{asild}, while providing timing guarantees as well as plug and play capability \cite{Sommer2013}.
Furthermore, the platform provides error detection and failure handling mechanisms on top of a real-time operating system and scheduler.
The research vehicle set up within RACE was equipped with many prototypical components to demonstrate the capability of a central platform computer and its middleware \cite{Buechel2015}.
Among these is a Steer-by-Wire system without mechanical fall-back, a prototypical intelligent brake actuator and non-production wheel hub motors.
The sensor set on the other hand is limited, cf. \refTable{tab:hardware}.
It could be demonstrated that the E/E architecture meets the requirements for a future homologation as a fail-operational system, which is mandatory for SAE Level 5 automation.
Nevertheless, the RACE vehicle demonstrator has no legal allowance to drive on public roads in Germany and is therefore not suited for autonomous driving algorithm design.

\textsl{fortuna}, on the other hand, shall serve as a platform for automated driving algorithm development, rapid prototyping and testing in real-world traffic scenarios, while always including a safety driver.
Hence it needs to meet production vehicle standards for the manual driving functions, which build the fall-back for ensuring road safety. 
Furthermore, all safety-critical hardware parts are certified production vehicle components.

\begin{table*}[tb]
	\centering
	\vspace{0.15cm}
	\caption{Software stack characteristics of the discussed vehicle platforms.}
	\scriptsize
	\ra{1.3}
	\begin{tabularx}{\textwidth}{p{1.5cm}p{1.5cm}p{1.5cm}p{1.5cm}p{1.5cm}p{1.5cm}p{1.5cm}p{1.5cm}p{1.5cm}}\toprule
 & VaMP \ignorecitefornumbering{\cite{Thomanek1996}} & Junior \ignorecitefornumbering{\cite{Montemerlo2008}} & Boss \ignorecitefornumbering{\cite{Urmson2008}} & Bertha \ignorecitefornumbering{\cite{Ziegler2014a}} & RACE \ignorecitefornumbering{\cite{Buechel2015}} & Halmstad \ignorecitefornumbering{\cite{Aramrattana2018}} & Bertha \ignorecitefornumbering{\cite{Tas2018}} & Apollo \ignorecitefornumbering{\cite{Baidu2017}}\\
 \midrule
Application & German Highway & \multicolumn{2}{c}{\gls{darpa} Urban Challenge} & German Rural & Parking & \multicolumn{2}{c}{Cooperative Driving Challenge} & Various\\
 \midrule
Licensing & proprietary & partly open & proprietary & proprietary & proprietary & proprietary & proprietary & open  \\
Middleware & ? & publish/ subscribe IPC & publish/ subscribe \gls{ipc} & ? & RACE RTE & LCM & \gls{ros} & Cyber RT \\
OS on PC & ? & Linux & ? & ? & PikeOS & Linux & Linux & Linux \\
Functional Safety & None & Watchdog module & Error Recovery & ? & supporting \gls{asild} & Trust System & ? & System Health Monitor\\
Controller on & ? & PC & ? &  ? & RACE DDC & MicroAutobox & realtime onboard comp. & PC \\
\bottomrule
\end{tabularx}

	\label{tab:software}
\end{table*}

\subsection{\textsl{fortuna} Compared to Other Research Vehicles}
\label{sec:related_work:other}

In this section, we discuss former research vehicles and compare their hardware and software setup to \textsl{fortuna}. 
We chose to include the vehicles of the two winning teams of the \gls{darpa} Urban Challenge and the \gls{gcdc}, respectively, as these competitions presented two milestones in autonomous driving research. 
\refTable{tab:hardware} depicts the hardware evolution of research prototypes in the past 15 years. We start by comparing the sensor setups. 

\subsubsection{Sensors}
VaMP, developed by \citet{Thomanek1996} in the 1990s, was limited to \gls{acc} and \gls{lc} applications on highways due to reduced sensor capabilities with only four cameras and computation hardware of approximately 50 processing units.
Fifteen years later, a successful choice for teams in the \gls{darpa} Urban Challenge was a fusion of high-end Lidar and Radar sensor data. 
However, back then perception systems could not rely on GPU-based acceleration and deep neural networks.
Since then, Radar and Lidar sensors for detection and camera-based systems for classification have been used in urban environments.
Recent advances in multi-sensor data fusion using machine learning methods have encouraged to use more advanced sensors and setups. 
With the Apollo reference vehicle \cite{Baidu2017} various scenarios have been demonstrated using different sensor setups.
The \gls{darpa} Urban Challenge established high-precision \gls{gps} as a standard. Since 2016, systems with \gls{rtk} have benefited from increased localization accuracy.

\subsubsection{Communication Interfaces}
Before 2016, research platforms did not focus on inter-vehicle cooperation or communication devices.
The first \gls{gcdc} \cite{VanNunen2012} in 2016 confronted researchers with cooperative \gls{acc} scenarios and introduced a V2V communication protocol. This challenge made appropriate communication devices necessary in the participating platforms. As fusing self-perceived sensor data and \gls{v2v} data proved to be challenging, \citet{VanNunen2012} selected to use the communication interface and a Radar, whereas \citet{Tas2018} selected to use the \gls{v2x} inputs only.
Similar competitors participated in the second \gls{gcdc} \cite{Aramrattana2018}, which also included lateral maneuvers with the need for communication and cooperation.

\subsubsection{Control}
Focusing on the hardware for control purposes, we observe no clear tendency towards a separation of control algorithms onto \gls{rtcomp} with vehicle bus access and soft real-time PC hardware (see Table~\ref{tab:software}).
Since using real-time computing platforms increases robustness to software crashes or system failures and ensures safe communication with the production vehicle hardware, we employ such a setup.

\subsubsection{Middleware}
Nearly all research vehicles rely on a customized middleware layer between the specific hardware and software modules. A comparison is shown in \refTable{tab:software}.
The open source \gls{ros} has evolved into a popular middleware for autonomous driving prototypes.
At the time of the \gls{darpa} Urban Challenge in 2007, \gls{ros} had not been released yet.
The winning teams implemented a network-based publish/subscribe \gls{ipc} system, which also formed the basis of \gls{ros}.
Although no information was provided on the software architecture used for the Bertha Benz Memorial Drive \cite{Ziegler2014a}, Bertha was running \gls{ros} during the \gls{darpa} Grand Cooperative Challenge in 2016 \cite{Tas2018}.
Baidu based the first versions of Apollo on an extended \gls{ros} by a downwards-compatible message protocol based on Google Protocol Buffers\footnote{\url{https://developers.google.com/protocol-buffers/}} and decentralized node management.
Starting from version 3.5, \gls{ros} was replaced by Cyber RT, a custom middleware that claims to be more performant and easier to use.
With Lightweight Communication and Marshalling (LCM) \cite{Huang2010} serving as an alternative, no standard has evolved until now.

Considerations of functional safety are commonly neglected with research vehicles. 
Nevertheless, watchdogs and sanity checks usually handle algorithm and system errors.

\section{Hardware Setup}
\label{sec:hardware}

\begin{figure}[tb]
	\centering
	\includegraphics[width=\linewidth]{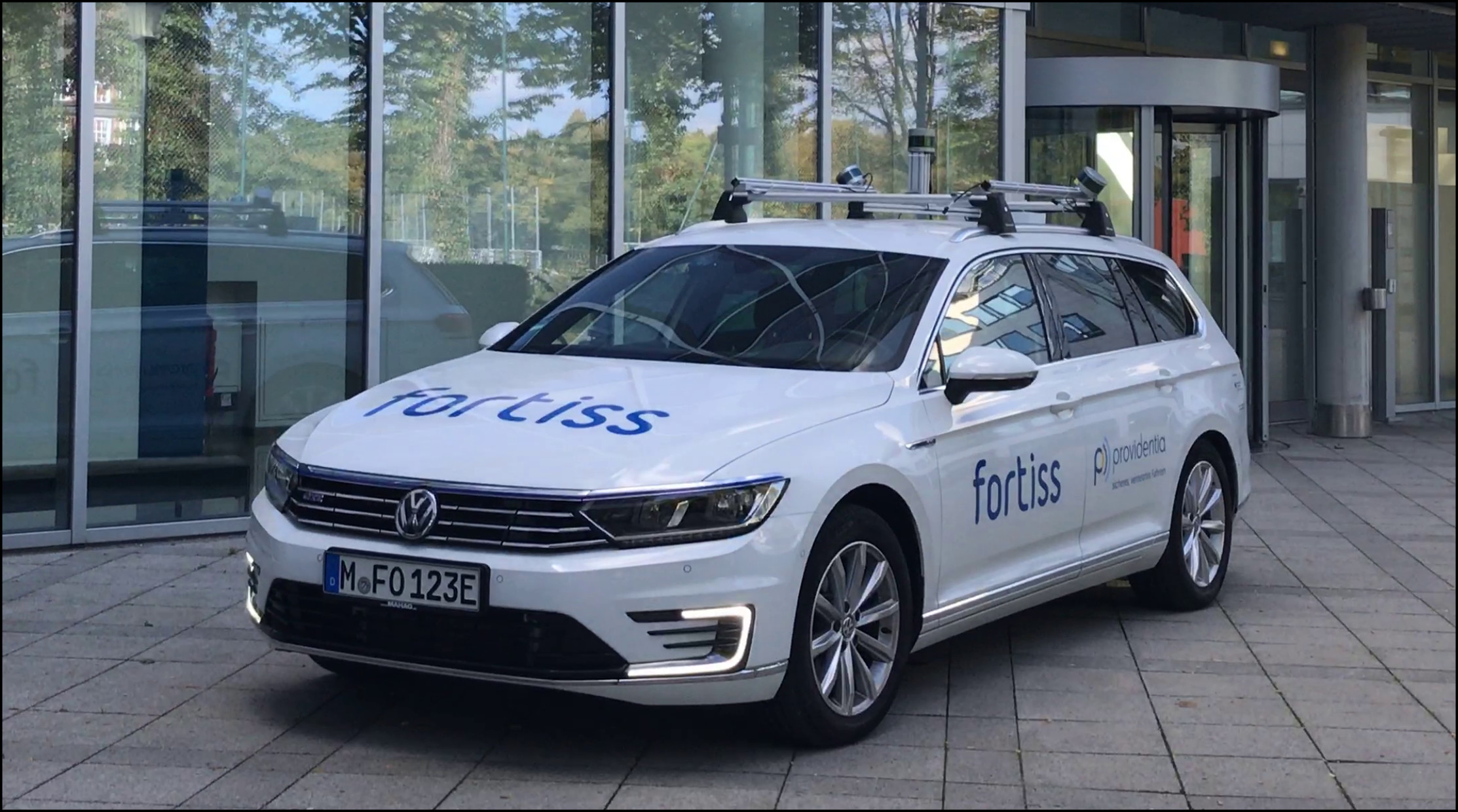}
	\caption{The \textsl{fortuna} autonomous driving vehicle demonstrator: a modified VW Passat with Lidar sensors and antennas on the roof rack, additional cameras inside the vehicle and additional radar sensors integrated in the bumpers.}
	\label{fig:fortuna_foto}
\end{figure}

To meet the requirements of rapid driving algorithm prototyping, we chose to modify a production vehicle with non-production-vehicle hardware as well as providing access to production sensors and actuators.
We refitted a 2018 Volkswagen Passat Variant GTE (see \refFigure{fig:fortuna_foto}).
This setup leverages the need for innovative and powerful hardware and the need for a safe and reliable base hardware setup.
We did not aim for a hardware setup of a production vehicle in terms of redundancy or power consumption. 

\begin{figure}[tb]
	\centering
	\vspace{0.15cm}
	\includegraphics[width=\linewidth]{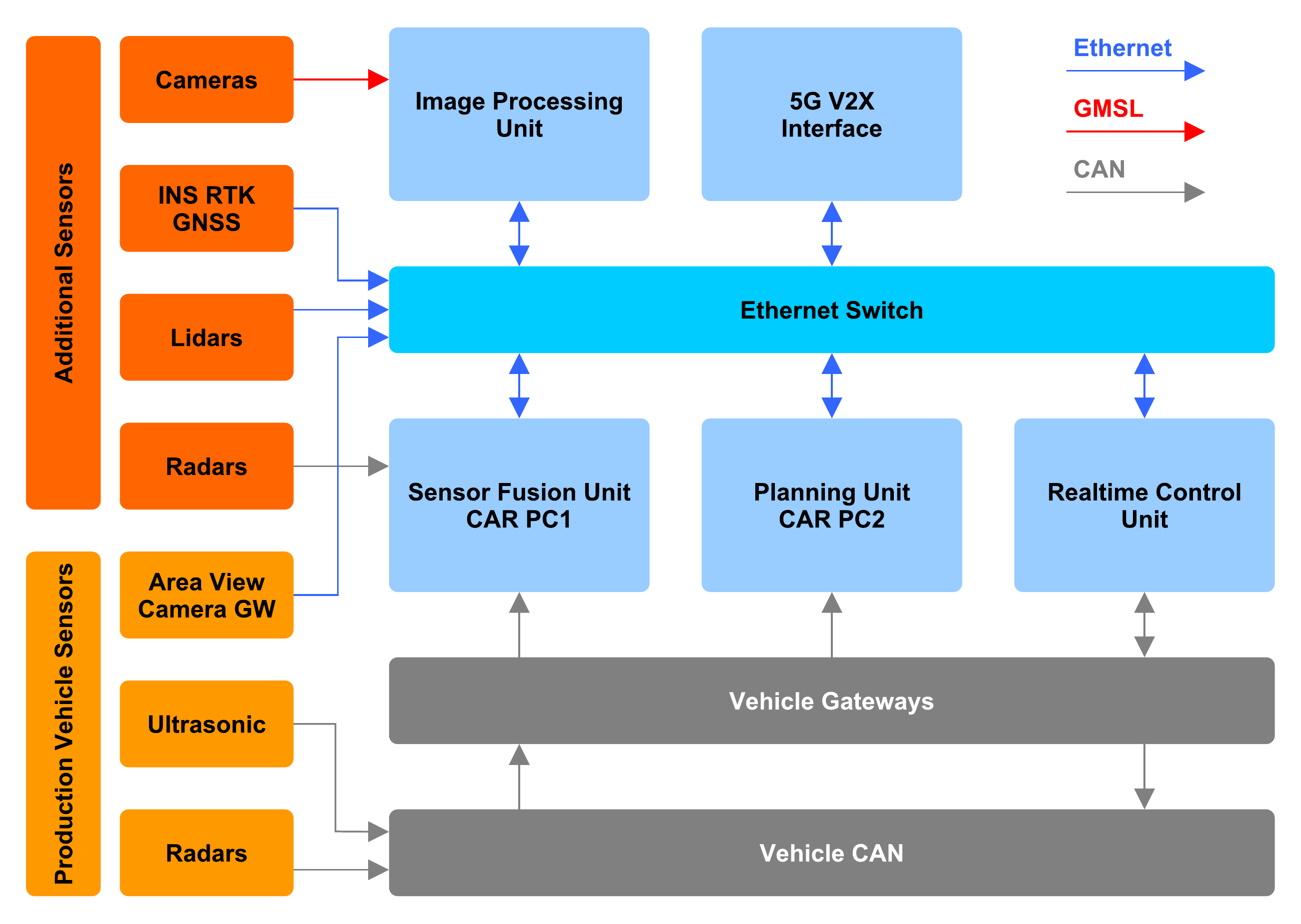}
	\caption{Schematic overview of the hardware setup and the interfaces between the components.}
	\label{fig:fortuna_HW}
\end{figure}

\begin{figure}[tb]
	\centering
	\vspace{0.15cm}
	\includegraphics[width=\linewidth]{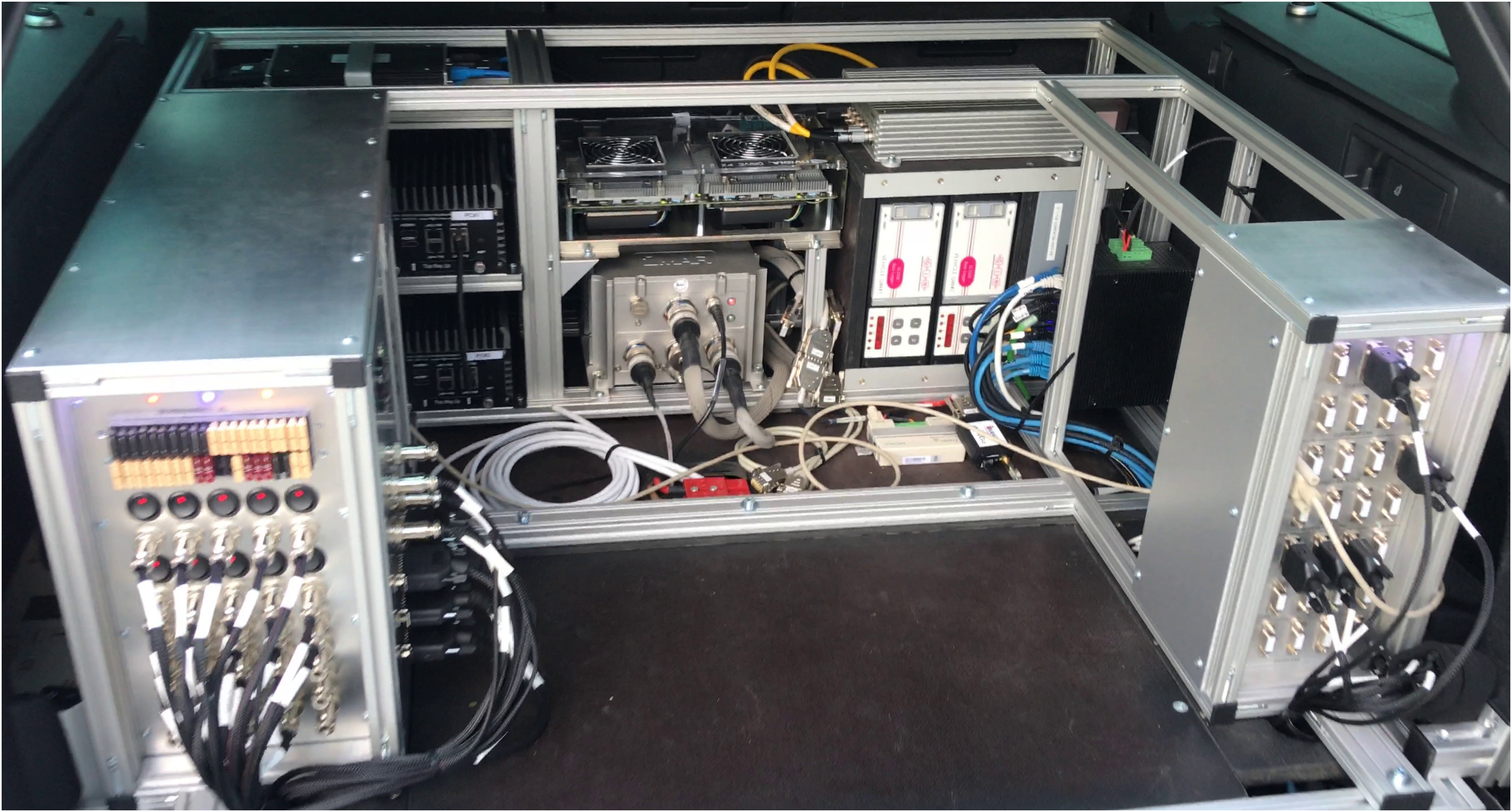}
	\caption{The additional hardware installed in the trunk showing (clockwise) the power supply, the two car PCs, the Drive PX 2, the iNat FSSG, the Micro Autobox, the CAN gateways, the Ethernet switch, and the CAN patch panel.}
	\label{fig:trunk}
\end{figure}

An architecture overview of \textsl{fortuna}'s additional hardware is depicted in \refFigure{fig:fortuna_HW}. \refFigure{fig:trunk} provides an impression of the installed setup in the trunk.
The modifications include additional sensors, interfaces to access production vehicle bus networks and four computers connected via an industrial gigabit Ethernet switch splitting the traffic into several virtual networks. 
\begin{itemize}
	\item One industry standard real-time rapid-prototyping control unit (a dSpace Micro Autobox II) with an IBM Power PC 900MHz CPU and 16MB RAM for control algorithms with various low-latency hardware interfaces (including CAN) running a real-time operating system. We run the low-level trajectory control on this computer as described in \Secref{sec:controller}.
	\item Two PCs with Intel i7 3.4GHz quad-core CPU and 32GB RAM for sensor data processing, motion planning, human-machine interfaces and further software components running Ubuntu Linux. These PCs run the driving stack as described in \Secref{sec:software}.
	\item One Nvidia Drive PX 2 AutoChauffeur with two Pascal GPUs running Nvidia Drive Works for accessing camera images. The platform is mainly used for vision-based perception and neural network inference as sketched in \Secref{sec:perception}.
\end{itemize}

The hardware setup includes a 12V backup battery with additional power management and a connection to the high-voltage system of the production vehicle.
A prototype cellular 5G interface realizes \gls{v2x} connectivity.

Proprietary gateways enable access to the CAN buses of the production vehicle, which allows reading sensor and vehicle information. 
Write access enables automated driving through steer-by-wire.
The production vehicle sensor data include object lists detected by the radar sensors of the \gls{acc} system and kinematic vehicle state information.
Also, raw data from the ultrasonic sensors and camera images from the Area View surround view cameras are available.

More in detail, the vehicle is equipped with the following additional sensors, cf. \refFigure{fig:fortuna_foto}, allowing a 360$^\circ$ \gls{fov} avoiding blind spots.
\begin{itemize}
	\item Three Velodyne Lidars: one VLP-32C with 32 layers in a central horizontal position on the rooftop and two VLP-16 with 16 layers at each side of the vehicle roof, inclined to scan the areas at each side of the vehicle. We regard Lidar sensors as mandatory for automated driving above SAE Level 3. The setup was chosen to provide a sound point cloud density in combination with a sufficiently broad sensor range for various scenarios.
	\item Five Sekonix cameras: two front-facing cameras, one with 60$^\circ$ \gls{fov}, and one with 120$^\circ$ \gls{fov}, one camera to each side and one rear camera, all with 120$^\circ$ \gls{fov}
	\item Four Smartmicro UMRR-146 radars: two facing forwards and two backward, integrated into the bumpers with access to raw sensor data
	\item \gls{ins}: iMAR iNAT FSSG-1, a fiber optic gyro (FOG) based \gls{ins} supporting \gls{rtk} with integrated \gls{gnss} receiver offering a localization precision of up to 2cm. The device can serve as a positioning unit and also as high precision reference localization for algorithm validation. As we consider a very high and reproducible localization measurement as essential for benchmarking autonomous driving functions, we decided on this industry-standard but non-automotive production grade device. 
\end{itemize}

Key switches are installed for safety reasons and allow to power and enable the reading access measurement system and to enable writing CAN access for longitudinal and lateral control.
An emergency shutdown button allows the safety driver to return to a production vehicle mode.

In contrast to the Apollo reference vehicle \cite{Baidu2017} or the Bertha vehicle \cite{Tas2018}, we chose to equip the vehicle with more than one computer to separate functionality. 
This setup comes at the price of network communication overhead and necessary design decisions on how to connect sensors and devices. 
For example, we connect the 360$^\circ$ \gls{fov} camera setup to the Drive PX 2 hardware via \gls{gmsl}.
This wiring hinders us from running image-based perception algorithms on one of the PCs.

\section{Driving Stack}
\label{sec:software}
We chose to base the software setup on the Apollo driving stack developed and maintained by Baidu \cite{Baidu2017}. 
Baidu claims that the stack contains all necessary modules for SAE Level 5 autonomy \cite{Baidu2017}. 
When starting this work, we used version 2.5, the newest at that time. 
As of now, we migrated our modules to version 3.5.
The stack has a very modular structure and brings its custom-built middleware to exchange information. 
Since the Apollo stack is embedded in a growing open-source community and many companies have joined the Apollo board, we chose to use Apollo over other open-source stacks such as Autoware \cite{Kato2018} or the Junior Driving Stack \cite{Montemerlo2008}.

For a comprehensive description of the Apollo software design and architecture, the reader is referred to the documentation in the open source repository\footnote{\url{https://github.com/ApolloAuto/apollo}}.
This section will focus on the extensions we needed to implement to run Apollo on our vehicle. 

To adapt Apollo and to run it on our research vehicle, we focused on modifying the localization, perception and controller modules. Additionally, we developed several adapters to connect the vehicle hardware with the Apollo stack. 
However, we aimed to introduce minimal changes to the Apollo stack to maintain the original functionality. Furthermore, we developed additional modules to verify the functionality of the stack on the research vehicle, such as a mocked localization and a mocked planner. We are thus able to send trajectories with various lengths, speeds and steering angles, making them an ideal validation tool.

Since we use different hardware for localization than the Apollo reference vehicle, we had to develop a customized localization adapter. 
As the INS/GNSS provides highly accurate and consistent measurements, we decided to feed an already filtered position into the stack.
However, we did not change the localization module to not lose any existing functionalities such as a watchdog that detects irregularities.

Our sensor setup is different from the recommended Apollo hardware-setup, and we re-implemented parts of the perception pipeline.
We describe these adaptions in the perception modules and adapters in greater detail in \Secref{sec:perception}.

We chose to replace the controller used by Apollo and run it on a separated real-time platform instead. 
This separation facilitates a higher level of safety by separating control tasks in real-time execution for other software applications on different hardware platforms. 
Therefore, we developed an adapter communicating between the Apollo stack and the real-time system along with a trajectory tracking controller suitable for real-time execution.
This software architecture will be described in greater detail in \Secref{sec:controller}.
We observed that this separation yields more stable system performance and enables us to optimally use the different benefits of the modular computing hardware setup.
The development of the autonomous function stack as a mixed-criticality system on a single computer is of high scientific relevance but addresses other research aspects than driving function development. 

In summary, Apollo provides prebuilt functionality and a well-structured code-base.
For the stack to run on our research vehicle, we had to change drivers, add adapters and customize modules.

\section{Perception and Sensor Setup}
\label{sec:perception}

\begin{figure}[t]
	\vspace{0.15cm}
	\includegraphics[width=\linewidth]{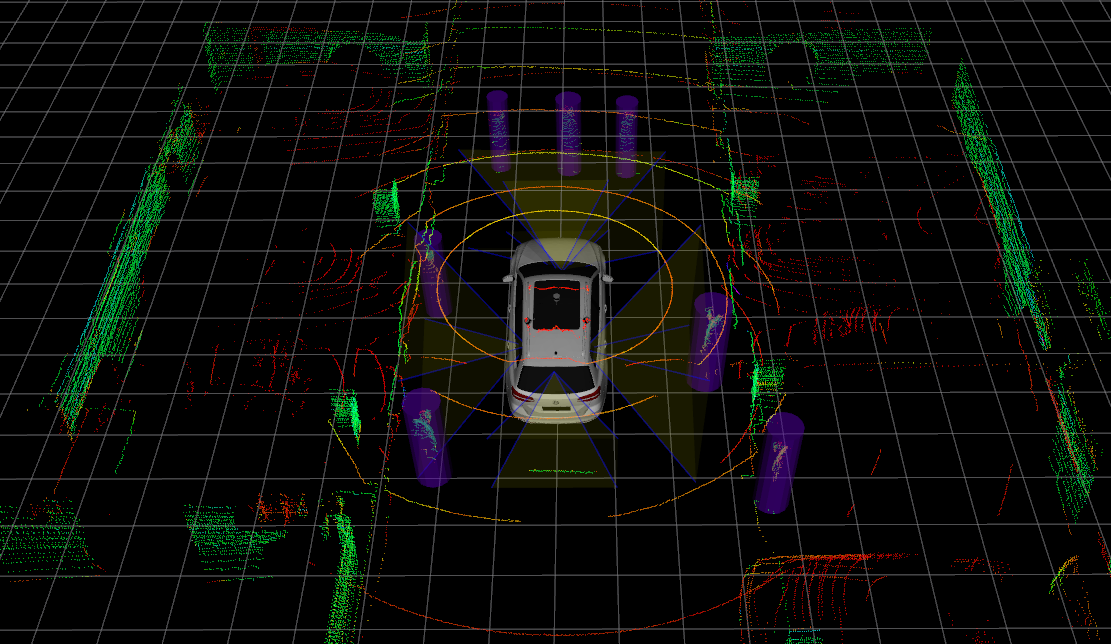}
	\caption{Lidar pointcloud with 3D pedestrian detections showing the perception components.}
	\label{fig:perception}
\end{figure}

This section describes how we calibrated the multi-sensor setup, integrated it into Apollo and further how we implemented a basic sensor fusion algorithm.
An example is shown in \refFigure{fig:perception}.

Up to this point, we calibrated all five Sekonix cameras and all three Velodyne Lidars towards a base link coordinate system which is located in the center of the rear axis. 
The section will conclude with a short discussion on the integration of the custom sensor setup into the Apollo framework. 
To calibrate the sensors we calculated the position of the central Lidar followed by a semi-manual calibration of each camera and the side-facing Lidars with respect to the central Lidar.

\subsection{Ground Plane Estimation}
For estimating the ground plane, it is essential to handle outliers. 
Especially point clouds generated from Lidar sensors are sensitive to distance and angle of incidence of emitted rays.
This explains why ground points further away from the sensor suffer from higher noise as the incidence angle gets sharper. 
To tackle this problem, we implemented a ground plane estimation algorithm based on \gls{ransac} \cite{Fischler1981}.
The model needed for the estimation is a plane which consists of 4 degrees of freedom
$\vec{n} \vec{x}^T + d = 0$
with the normal vector $\vec{n} \in \mathbb{R}^3$ and an offset $d \in \mathbb{R}$ of the plain. 
A point $\vec{x} \in \mathbb{R}^3$ lies on the plane if the equation evaluates to true.

On each \gls{ransac} iteration, at least three points are randomly selected to form a plane. 
All the remaining points are evaluated on the plane's equation by thresholding the plane-point distance.
The points on the plane are saved in the consensus set. 
After convergence, all points in the final set are used to estimate the final ground plane. 

This algorithm assumes that a large portion of the point cloud is the ground plane. Therefore, wide and flat locations -- such as parking lots -- are preferred.

\subsection{Semi-Manual Lidar-to-Image and Lidar-to-Lidar Calibration}
For calibrating each camera to the vehicle base link coordinate system, we estimated the position of each camera separately towards the central Lidar following the work of \citet{Dhall2017}\footnote{Implementation of \url{https://github.com/agarwa65/lidar_camera_calibration}}.
We selected more than six 3D-2D point correspondences manually in the point cloud and the corresponding image.
These points are then used to solve for the projection matrix $\mathbf{P} = [\mathbf{R} | \vec{t}]$, in which the rotation matrix $\mathbf{R} \in \mathbb{R}^{3\times 3}$ is represented by its yaw, pitch, and roll angles.
The vector $\vec{t} \in \mathbb{R}^3$ is the resulting translation from the central Lidar coordinate system to the camera origin. 
The solution for the overdetermined linear equation system was then estimated by a Sequential Least Squares Programming optimization algorithm.

We calibrated the two side-facing Lidars by using the \gls{icp} \cite{Besl1992} algorithm to estimate the six \gls{dof} transformation between the point clouds of each side-facing Lidar and the central Lidar, with a rough initial estimate. To further improve the calibration quality manual fine tuning was performed.

\begin{figure*}[t]
	\centering
	\vspace{0.15cm}
	\tiny
	\def\svgwidth{\textwidth}
	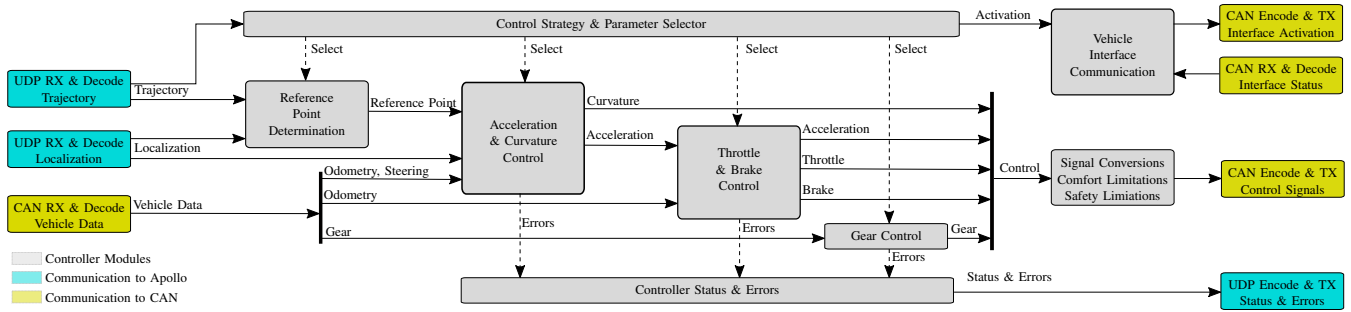
	\caption{Architecture overview of the trajectory tracking controller with the longitudinal/lateral vehicle control flow from the left to the right and the module parametrization and error handling flow from top to bottom.}
	\label{fig:controller}
\end{figure*}

\subsection{Perception Pipeline and Apollo Integration Challenges}
As a first integration step, we set up a fusion of camera detection and Lidar point clouds for object detection based on frustums \cite{Qi2017} independent of Apollo using standard \gls{ros} making use of the 360$^\circ$ \gls{fov} of camera and lidar. 
A state-of-the-art deep learning based object detector \cite{Huang2017} was used to detect pedestrians in all cameras. 
The bounding box was then projected into 3D Euclidean space resulting in four lines representing the corners of the 2D bounding box. 
The area inside those lines is called frustum. 
All points outside this frustum were removed, and the resulting points were clustered using DBSCAN \cite{Ester1996}. 
For simplicity, we assume that the target object is not occluded and consists of sufficiently many points so that the closest cluster to the ego-vehicle can be chosen as a detection candidate. 
Afterward, a cylinder is fitted to the cluster leading to the detection.
\refFigure{fig:perception} visualizes the algorithm.

We selected \textsl{fortuna's} sensor set before Apollo was released and also with highway scenarios in focus.
It is thus independent of the Apollo perception pipeline inputs.
We faced several challenges while integrating the sensor set when we started using Apollo 2.5.
With more recent versions (3.5 as of now) the integration barriers decreased, and mainly configuration work was necessary. 

Since the cameras are connected to the Drive PX 2, Apollo's perception components need to be evaluated directly on the device to avoid costly forwarding of raw camera data to another PC. 
We consider this future work as well as the usage of other sensors mentioned in \Secref{sec:hardware}. 
As a proof of concept connect a standard off-the-shelf USB camera to the car PC along with the central Velodyne-32 Lidar sensor and run the Apollo perception module out of the box.

\section{Trajectory Control}
\label{sec:controller}

This section describes how we implemented and tested our trajectory tracking controller.
As main contributions, we consider the description of the communication setup between the non-real-time trajectory planner, the real-time trajectory controller and the vehicle control units (cf. \Secref{sec:controller:interfaces:planner} and \Secref{sec:controller:interfaces:vehicle}).
Also, we show how we extended a trajectory control algorithm from literature. \Secref{sec:controller:algo} also states the architecture of the component.
Furthermore, we briefly outline our test and analysis tooling in \Secref{sec:controller:test}.

\subsection{Control Algorithm}
\label{sec:controller:algo}
The control algorithm is based on the work of \citet{Werling2010} and shown as a block diagram in the middle part of \refFigure{fig:controller}. 
Based on the driving situation we either apply a full trajectory or a path tracking mode. 
The suitable control strategy and parametrization is selected using the received trajectory. 

In the trajectory tracking case, we interpolate using the time on the given trajectory. 
As a common time base for the trajectory planner and the controller, we use the localization time signal. 
In the path tracking mode, the point with the closest Euclidean distance to the current vehicle pose on the trajectory is extracted as the reference point. 
The trajectory planner has to ensure that a sufficient backward horizon of the trajectory is available to guarantee a valid result of this interpolation.

A full trajectory tracking is especially valuable when driving on longer road segments with sufficient time and space to cope with sensor and actuator errors. 
When maneuvering, time aspects are of minor importance, and the situation is less dynamic. Trajectory tracking can also yield suboptimal final poses.
The errors and their derivatives are extracted from the tracking point in a Frenet reference frame. 
Using input substitution and backstepping asymptotic stability of the control law can be proven \cite{Werling2010}.

For the sake of driving smoothness and planner error tolerance, we limit the absolute values and rates of all control signals. 

A separated software module, implemented as finite state automaton, activates and parametrizes the different controller components based on the received trajectory and a host PC HMI component (omitted in \refFigure{fig:controller}).
With this, we achieve a full separation of control algorithm code from functional execution logic.

We perform basic consistency checking of each trajectory and localization signal in terms of data and time validity. Furthermore, we detect actuator failures and vehicle interface errors. 
In case of an error, the control is handed over to the safety driver.
More sophisticated errors like a bad tracking quality are expected to be treated by higher level components.

\subsection{Trajectory Planner Communication Interfaces}
\label{sec:controller:interfaces:planner}
The control algorithm is running on a rapid prototyping hard real-time computation platform whereas a PC-like hardware executes the trajectory generation components in a soft real-time context. 
As the trajectory is available to the controller over a particular horizon, no real-time communication between planner and controllers has to be implemented. 
Consequently, delays or packet loss in communication become acceptable. 
Also, short planner computation delays still result in smooth motions.
In case of a planner software failure, the controller can still evaluate the last valid trajectory and can trigger an emergency stop or hand over the control to the safety driver.

The Apollo trajectory planner outputs a collision-free trajectory and transfers a sampled representation on a defined horizon to the trajectory controller. 
The controller performs no more collision checks allowing it to run at a high frequency.

\refFigure{fig:controller} depicts the interfaces of the controller and the communication channels.
The communication to the trajectory planner and the localization module is realized using Ethernet with a UDP protocol. With the interpolation method described in \Secref{sec:controller:algo}, no assumptions on the time or spatial distance of trajectory points are required. As an encoding format of the UDP messages, we use the protocol buffer definitions for trajectory and localization from Apollo.

\subsection{Vehicle Communication Interfaces and Execution Platform}
\label{sec:controller:interfaces:vehicle}
In contrast to the Apollo reference vehicle, we separate the computers for planning, perception and other driving functions from the closed loop vehicle control. 
That way, in case of timing problems on the computers or Ethernet network outtakes, we still can keep up the control loop on a short horizon. 
The trajectory controller is executed on a dSpace Micro Autobox II having a cycle time of 10\,ms. Only this real-time hardware holds access to the vehicle controls.

The controller does not directly influence the vehicle actuators but computes high-level control values like an acceleration and steering wheel angle command.
To control the vehicle motion a subsequent vehicle gateway control unit uses these high-level control values and actuates the production vehicle's control unit interfaces. The main interfaces are
\begin{itemize}
	\item the acceleration interface from the production vehicle's automatic cruise control,
	\item the steering wheel angle interface from the production vehicle's park assistance system.
\end{itemize}
For low speed and maneuvering scenarios with reverse driving segments, the gear selection, throttle, and brake are actuated directly.
No modification of any production control units is necessary. The implementation of the low-level control interfaces and the actuation of the production vehicle control units are realized on private CAN buses. These are proprietary and out of the scope of this work.
\subsection{Implementation, Test, and Analysis}
\label{sec:controller:test}
We modeled the controller and the CAN and Ethernet interfaces in Mathworks Simulink with certain code parts embedded as C-Code S-Functions.

For a seamless module test without the real-time hardware, we execute the model on a development PC. Mocking the same interfaces as the vehicle CAN bus enables us to run open or closed loop tests with the controller running in Simulink on a computer while receiving trajectory and localization from Apollo or recorded data.
Furthermore, code generation to implement a virtual controller as Apollo node is planned to automate the test setup further.

To debug errors, it has been proven valuable to record all input signals and internal controller states using the toolchains offered of the different platforms \cite{Minnerup2016}. A unified analysis and plotting framework has been implemented to analyze the recorded data.

\addtolength{\textheight}{-2cm}

\section{Conclusion}
\label{sec:conclusion}

In this work, we pointed out the challenges, pitfalls, and lessons learned we encountered while integrating and running the open-source driving stack Apollo on our research vehicle \textsl{fortuna}. The research platform is an ideal basis for further research on functional autonomous driving software components, especially with the legal allowance to drive on German roads.

We integrated our vehicle's hardware into the Apollo software stack by adding novel adapters, such as for the localization and vehicle control. 
Decoupling the controller from the other software stack components enables a safe control -- also in case of prototype software or hardware failures.

However, we experienced notable engineering barriers while integrating Apollo with our vehicle hardware and sensor set.
We separated the perception pipeline and sensor calibration from the Apollo stack and tailored it to our sensor set.
Also our non-centralized computation hardware led to modifications in the driving stack. 
Nevertheless, Apollo proved to be a good choice to base our driving functions on due to its modularity and modern software design that made the integration of our custom components possible.

Future work will focus on a detailed evaluation on how Apollo performed with our hardware setup, including the discussion if this vehicle setup was a reasonable choice.

One of our next research goals is to modify the existing open source driving software stack to enforce a reliable vehicle behavior and the source code contribution to the community. 
We plan to address limitations of the ISO26262 norm for functional safety regarding autonomous driving.

\printbibliography

\end{document}